# Title: GMPilot: an expert AI Agent for FDA cGMP compliance

**Author**：Xiaohan Wang[1,†]; Nan Zhang[1,†]; Sulene Han[2,†]; Keguang Tang[1]; Lei Xu[1]; Zhiping Li[1,3,*]; Xiue (Sue) Liu[4,*]; Xiaomei Han[1,*].

**Institute:**

[1]Aelion AI Inc.

[2]H&N Consulting

[3]Department of Medicine, Johns Hopkins University, Baltimore, Maryland.

[4]Gilead Sciences Inc.

[†] These authors contributed equally to this work.

[*]**Corresponding authors:**

Xiaomei Han. Email: xiaomei.han@aelion-ai.com

**Abstract:**

The pharmaceutical industry is facing challenges with quality management such as high costs of compliance, slow responses and disjointed knowledge. This paper presents GMPilot, a domain-specific AI agent that is designed to support FDA cGMP compliance. GMPilot is based on a curated knowledge base of regulations and historical inspection observations and uses Retrieval-Augmented Generation (RAG) and Reasoning-Acting (ReAct) frameworks to provide real-time and traceable decision support to the quality professionals. In a simulated inspection scenario, GMPilot shows how it can improve the responsiveness and professionalism of quality professionals by providing structured knowledge retrieval and verifiable regulatory and case-based support. Although GMPilot lacks in the aspect of regulatory scope and model interpretability, it is a viable avenue of improving quality management decision-making in the pharmaceutical sector using intelligent approaches and an example of specialized application of AI in highly regulated sectors.

# 1. Introduction

Artificial Intelligence (AI) and the recent advances in Large Language Models (LLMs) and generative AI provide new technological opportunities to streamline processes and decision-making in knowledge-intensive sectors [Raiaan et al. 2024; LeCun et al., 2015]. In pharmaceutical enterprises, one of the core functions is the Quality Management System (QMS) which ensures safety, efficacy and quality of products [Bhikadiya, 2024]. This important role however has two challenges. The regulatory expectations are rising, with great scrutiny from regulations such as FDA.

Examination of the latest FDA reports indicates that there is a definite tendency: despite the fact that the quantity of registered establishments stays constant, FDA inspection activity has grown by around 6% [FDA, 2023, 2024]. Combined with a policy change towards greater unannounced foreign inspections [FDA, 2024], this trend marks a new norm of increased inspection activities.

Internally, the application of quality management in the industry still depends a lot on personal experience, retrieval of fragmented documents and judgment under pressure. This model reveals systemic vulnerabilities, including delayed problem solving, inconsistent standards and knowledge gaps when faced with complex global supply chains, rapid production cycles and unexpected events.

These issues are especially acute in the environment of current Good Manufacturing Practice (cGMP) regulation and enforcement. cGMP regulates all aspects of product quality control, monitoring form raw materials, facility design to integrity of data, etc. in pharmaceutical companies for drug ingredients, drug products and medical devices [Bhikadiya, 2024]. Most of its principles are interpretative where it states what should be done and how to do it. Conventional workflows that were mainly reliant on Standard Operating Procedures (SOPs) and personal experience, are now finding it difficult to effectively meet this contemporary quality management requirement which demands high correlation of knowledge and immediate insight.

Although general-purpose AI assistants are convenient and have extensive knowledge systems, they lack the focus and guidance to retrieve hyper-specialized domain specific information required to make informed decisions in compliance domain. They therefore do not fulfill the non-negotiable criteria of data accuracy-contextual relevance and traceability in compliance domain [Lee et al., 2020; Zhu et al., 2025]. In this paper, we suggest and create GMPilot (https://gmpilot.co/), a highly specialized artificial intelligence agent that will be easily incorporated into the everyday activities of quality departments. It is intended to serve as an intelligent compliance professional with the FDA cGMP regulatory environment.

The main contributions of GMPilot are: (1) the suggestion of a specialized AI agent structure in the highly regulated pharmaceutical industry that is strongly integrating ReAct and RAG; (2) the building of a high-fidelity FDA cGMP compliance knowledge base, which is built on official documents (Form 483s and CFR regulations), with strict manual verification; (3) proving the usefulness of the system in improving response efficiency, decision professionalism and traceability of knowledge and demonstrating its capacity to optimize the existing workflow, using a simulated inspection scenario; and (4) systematically investigating the potential of GMPilot to restructure QMS workflow. Our hypothesis is that by designing GMPilot as proprietary knowledge base and specific architecture, it can offer more focused and dependable support to the pharmaceutical sector than general-purpose AI systems. Its design is described in the following sections and its application value is illustrated by a practical simulation and also discussed in depth its great importance as an AI agent to the healthy long-term development of pharmaceutical industry.

## 2. Related Work

The swift development of LLMs, which was enabled by the transformer architecture [Vaswani et al., 2017], has provided the underlying capability of such systems. The evolution of few-shot learning in GPT-3 to instruction-following ability in ChatGPT has led to a change in the AI direction to alignment, safety and practicality [Vaswani et al., 2017].

### 2.1 Domian-Adaptive Pre-trained Language Models

The use of AI tools in life sciences is well known, and substantial progress has been made in the fields such as protein modeling and medical Q&A systems [Raiaan et al., 2024, Kaddour et al., 2023]. One of the important relevant precedents is BioBERT [Lee et al., 2020], a language representation model that was further pre-trained on massive biomedical corpora, such as PubMed. BioBERT showed that further pre-training using high quality domain-specific corpora can be highly effective in improving accuracy of the model to comprehend and extract information tasks like named entity recognition and relation extraction. The success of BioBERT confirms the necessity of high-quality domain corpora to develop professional AI tools. GMPilot implements an analogous domain customization strategy but it dwells more on the comprehensive knowledge and systematic application of pharmaceutical regulatory and compliance texts.

### 2.2 AI Agents and Enterprise Compliance Systems

Over the past few years, as LLM agents and tool-calling have developed, several exploratory systems have been introduced with the goal of strongly embedding AI in enterprise everyday operations. Compliance Brain Assistant (CBA) is one of such studies, CBA is a conversational agentic AI system that has been specifically created to be used in enterprise compliance situations [Zhu et al., 2025]. Zhu et al. indicates that CBA can be many times more accurate than general-purpose AI in compliance tasks (average match rate increased from 41.7% to 83.7%) due to a specialized knowledge base and integration with enterprise tools. This also confirms the efficiency of the domain-specific knowledge base in constructing expert systems. Unlike the focus on the overall enterprise compliance, GMPilot is devoted to help on solving the specific complexities of pharmaceutical cGMP compliance. The present work will concentrate on integrating a domain-specific knowledge base of pharmaceutical compliance with an agent workflow based on ReAct framework to directly address the unique regulatory and operational issues of the pharmaceutical sector that requires highly traceability and real-time responsiveness.

## 3. The GMPilot AI Agent

GMPilot is designed as a smart core that will be integrated into the pharmaceutical enterprise quality management system and will serve as a 24/7 quality expert. Its main aim is to enable and empower the quality team and change the workflow of manual information retrieval to proactive quality management based on continuous intelligent support. The conventional QMS workflow involves an inspection trigger which leads to a manual search in PDF files and SOPs, which strongly depends on personal needs. The GMPilot-augmented workflow brings the following innovative enhancements: (1) As a real-time knowledge base, offering instant access to relevant regulatory information; (2) Audit/Inspection readiness: Combines thousands of historical Form 483s from FDA inspections and latest regulations, automatically creating customized, risk-based inspection checklists and preparation points based on the context of a specific production line or product; (3) Daily practice: Functions as an online consultant available 24/7, providing best practice guidance during batch record review, change control assessment, supplier quality evaluation and other routine operations, etc. (4) Guidance for Deviation, OOS (and other cGMP compliance related issues) management, such as: For un-planned deviation, investigation on the root cause, and followed by corrective and preventive action (CAPA). For planned deviation, explain the reasons.

### 3.1 System Architecture and Workflow

The technical implementation of GMPilot is designed as a unified AI agent, which emulates an experienced quality expert. The system is built upon the fundamental

ReAct (Reasoning and Acting) framework as its underlying paradigm, with context-sensitive prompt engineering dynamically orchestrating reasoning pathways and action selection, allowing the LLM to perform differentiated problem-solving behaviors specific to task situations [Yao et al., 2023].

The system is based on the ReAct paradigm, which involves iteratively Reasoning about a task, acting to gather information or perform a sub-task, and observing the results to inform the next step. Reasoning path, tool calling and analyzing observations are determined by user's initial input and contextual dynamic in system evolution.

The specific nature and complexity of this loop are determined in real - time by the initial user query and the evolving context of the system.

In the case of simple queries, the agent performs a brief loop: thinking that a direct retrieval is required, acting to retrieve the regulation and immediately synthesizing an answer (Figure 1). In multi-faceted tasks which are complex, the first prompt will set off a more advanced ReAct loop. The agent initially thinks to break down the task into sub-goals. It then proceeds to act on each of the sub-goals one after the other, retrieving and synthesizing information prior to proceeding to the next, in a single controlled thought process (Figure 1).

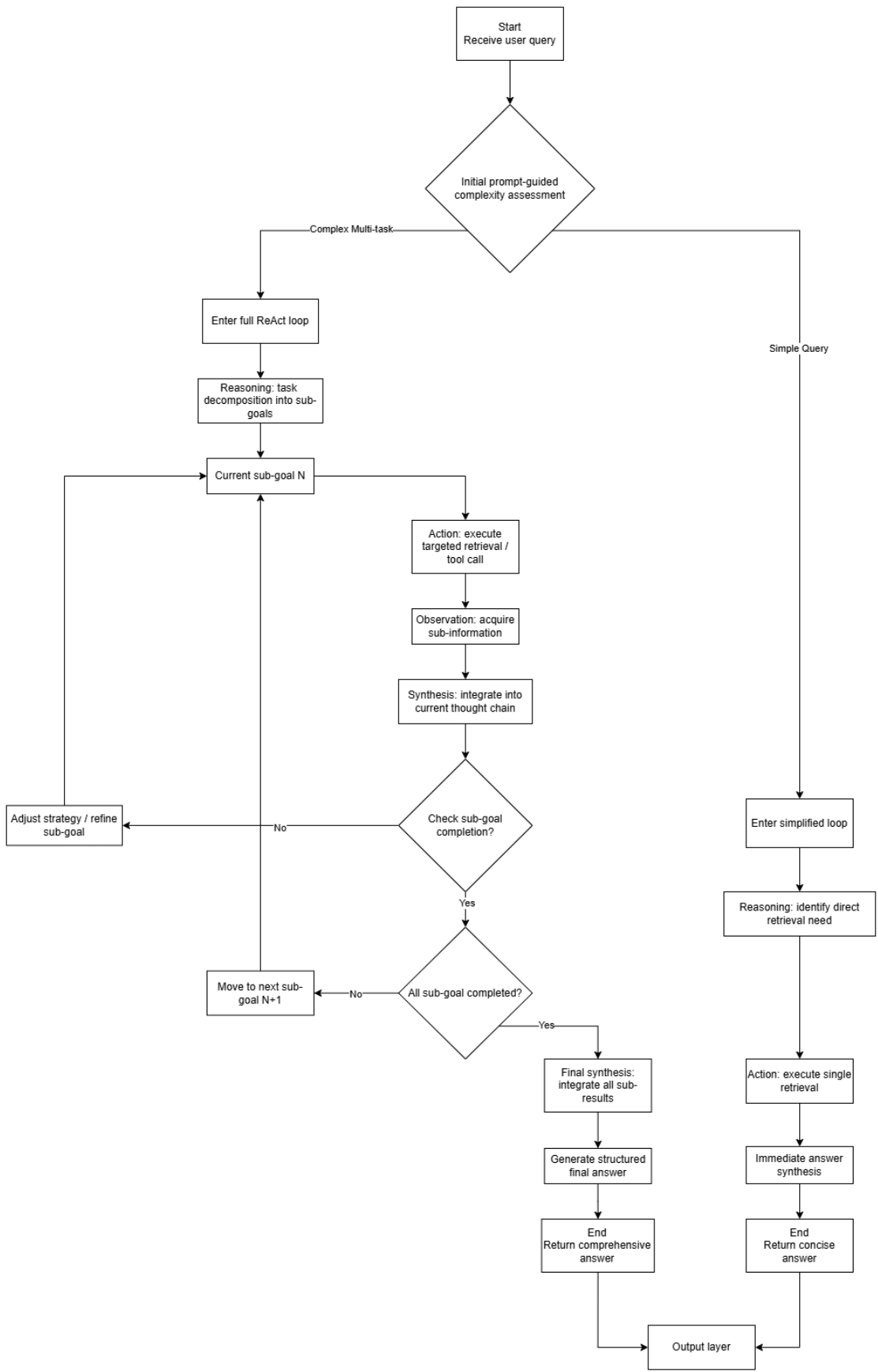

Figure 1: GMPilot Workflow

This agent system is prompt-based, which allows the system to be flexible in performing a broad spectrum of tasks, including simple Q&A and complex planning, with consistency and explainability in decision logic, and without the integration complexity and maintenance cost of multi-module systems.

## 3.2 Knowledge Base and Retrieval-Augmented Generation

The professional knowledge of the system is based on its specialized knowledge base that ensures the authority and domain proficiency of GMPilot. This paper builds and sustains a curated FDA cGMP compliance corpus with multi-sources and heterogeneous ground-truth data. It contains the text of cGMP regulations and an authenticated set of publicly available FDA Form 483 observations accessed through official sources. All the documents are subjected to intensive manual checks and regular updates to create a uniform high-fidelity compliance data asset.

At the time of the statistics cut-off, there are 1598 Form 483s in the dataset. To eliminate optical character recognition (OCR) noise that is present in unstructured PDF documents, every document will be manually checked character-by-character to determine text ground truth. This procedure broke down the content into 10891 high-precision and independent observation items. Moreover, we have also applied a manual confirmation-based entity disambiguation and alignment process to resolve inconsistencies in naming entities in the original records. This created firm and inspector relationships with distinctness between inspected entities and consolidated them into 775 unique firm groups and standardized inspector references into 1114 unique personnel groups. These preprocessing procedures of building ground-truth data were successful in removing statistical noise so that further quantitative analysis would be reliable. The strict human processing of underlying data quality cannot be matched by the open web data used by general-purpose AI tools which is the foundation of GMPilot quest to achieve high accuracy and relevance in the pharmaceutical regulatory compliance domain.

Besides, the dataset has a compliance mapping framework which is highly specific to the pharmaceutical industry and includes 60 FDA CFR Parts that are very pertinent to pharmaceutical regulation. This proprietary regulatory tree includes basic cGMP regulations (21 CFR Parts 210, 211) and full lifecycle compliance requirement, such as IND/NDA/ANDA submissions (Parts 312, 314), and biologics (Parts 600-680). This compliance mapping offers multidimensional compliance risk analysis and quantification.

This is a curated corpus that allows the retrieval-augmented generation (RAG) process to be targeted and seamlessly integrated into the ReAct cycle. Once the reasoning of the agent establishes that factual retrieval is necessary, it accesses this specialized database to retrieve the most relevant regulatory clauses and excerpts of real-world inspection results. The sourced documents are subsequently fed into the prompt, where the LLM will be directed to produce a regulation-aligned, accurate, and citation-supported response. This architecture guarantees that all outputs can be directly linked to official regulatory texts and practical precedent with accuracy being prioritized over generative creativity.

### 3.3 Design Rationale and Implementation Configuration

We selected an agent architecture with a single knowledge base to achieve consistency and eliminate the overhead of integrating various specialized modules. RAG mechanism was chosen because it gives a transparent connection between the generated advice and its source, which is an important requirement in pharmaceutical industry. The main idea behind the entire system architecture is to make the output content as practically and reliable as possible in the serious context of pharmaceutical compliance by building a manually verified, domain-specialized and high-fidelity knowledge base and combining it with the RAG mechanism and ReAct framework. This design goal is essentially unlike the general coverage and generative abilities that are sought after by general-purpose AI.

### 3.4 System Implementation and Technical Configuration

To achieve system, robustness, reproducibility and ability to process long texts in compliance situations, GMPilot is developed on the Dify orchestration framework with specific technical settings.

### 3.4.1 Base Model Configuration

Considering the intricacy of pharmaceutical compliance reasoning, we implemented a high-performance open-source LLM (Qwen3, 235B parameters) as the reasoning core. This specific base model and configuration are subject to evolution as high-language-model capabilities continue to advance. The model is set up with an ultra-long context window of 131,072 tokens, which allows it to process whole, complex regulatory documents and lengthy inspection reports in one pass, effectively avoiding logic loss due to context truncation.

### 3.4.2 Adaptive Data Chunking Strategy

In order to deal with the multi-source heterogeneous characteristics of knowledge base data, we adopted differentiated chunking strategies to balance semantic integrity and retrieval granularity: (1) Regulatory corpus: natural paragraphs are used as separators, with a chunk size of 1024 characters and an overlap rate of 5%. This setting is intended to maintain the autonomy of regulatory clauses and at the same time retain semantic relationships between neighboring clauses; (2) FDA 483 supplementary corpus: in order to preserve the contextual totality of historical inspection events (e.g. the relationship between observation items and particular backgrounds on production lines), we have applied a custom delimiter strategy, where chunk size is set at 40,000 characters (with no overlaps), ensuring each retrieval unit strictly corresponding a Form 483 observation and full retention of semantic relationships between observations, deficiency and production line. The approach takes advantage of the long-context feature of the base model, which makes sure that the model does not see the cases broken down into sentences but rather complete ones; (3) Expert Q&A corpus: uses a Q&A segmentation mode, whereby questions and their standard answers are retrieved as tightly coupled units, enhancing the directness of matching.

### 3.4.3 Hybrid Retrieval and Re-ranking Mechanism

To fill the matching gap between professional terminology (e.g. specific regulation codes) and semantic description, the system uses a hybrid search strategy which is based on keyword-based full-text search as well as vector-based semantic search. Results are fused and ranked through weight configuration, in order to meet the requirements of both exact term matching and semantic similarity matching. Then, the retrieved results are also subjected to secondary filtering by a Re-Rank model to further suppress hallucinations and enhance precision. We established a strict similarity threshold of 0.7 where we kept only top 2 most relevant content to feed into LLM. This process maximizes the suppression of irrelevant interference and ensures the output content is strictly based on Hi-Fi regulations. This less but better approach works well in minimizing noise interference in irrelevant reasoning within the model and making sure that the advice generated is strictly confined to high-confidence evidence.

### 4. Use Case: Simulating an Inspection Scenario

To concretely show the value of GMPilot as a quality expert, we simulate a scenario covering internal discovery to external response. The goal is to qualitatively analyze the system's workflow, output features and ways to improve existing workflows.

### 4.1 Simulated Scenario

Scenario: A serious aseptic control deficiency was found in a pharmaceutical industry production line, and it may triggers product recall and FDA re-inspection. The quality team had to start an investigation right away and get ready for the re-inspection.

User query input: for aseptic control deficiency, write a brief preparation and CAPA for the FDA re-inspection, if FDA re-inspection results are not satisfied, products recall maybe triggered.

### 4.2 GMPilot's Workflow and Output Analysis

Upon receiving the query, GMpilot initiates the ReAct loop:

Reasoning: determines the task to generate an inspection preparation brief and breaks it down into sub-goals: retrieve relevant regulations, retrieve historical similar cases and synthesize and generate structured recommendations.

Acting and observing: Sequentially accesses the knowledge base key clauses including 21 CFR 211.42 (Design and construction features), 211.46 (Ventilation, air filtration, air heating and cooling), 211.100 (Written procedures; deviations), 211.113 (Control of microbiological contamination) and recent FDA Form 483 observations about aseptic control deficient and related system control deficiencies.

Synthesizing response: Returns a structured dossier within minutes, containing:

a. Direct regulatory basis: Excerpts of relevant CFR regulations.
b. Historical precedents summary: A presentation of recent relevant FDA Form 483s finding lists with their specific deficiency descriptions.
c. Actionable inspection checklist: Based on regulations and cases, lists violation risk summaries and action items.

### 4.3 Value Discussion

This simulation demonstrates the GMPilot's workflow and its possible usefulness in compliance situations. The main features of the output of the system are high contextual relevance, traceability and actionability. More to the point, the content of its outputs and recommendations are directly connected with particular regulatory provisions and associated historical cases, creating a detailed and complete chain of evidence and is displayed in a structured form to be executed quickly.

The workflow transformation introduced by GMPilot is therefore primarily manifested in two ways: First, it reduces the critical time of information processing of preparation to inspection down to hours even days of traditional manual work to real-time and minute-based responses with a considerable improvement in efficiency. Second, it allows quality professionals to move their attention away from labor-intensive gathering of information towards high-value activities including risk analysis, strategy formulation and implementation monitoring through instant and verifiable contextual information. Thereby facilitating a shift in work patterns between passive response and proactive quality assurance.

## 5. Limitations

The main weakness of GMPilot is its jurisdictional scope, as it is currently designed to apply only to FDA cGMP and cannot be applied directly without adaptation to operations under the European Medicines Agency (EMA), National Medical Products Administration (NMPA), or other authorities.

Moreover, being an LLM-based system, it has intrinsic limitations. It is limited in its performance by the knowledge base; new situations that are not known to have happened before can test their correctness. Its synthesis is black-box and may make it difficult to see why certain sources should be prioritized over others, requiring continued critical supervision on the part of users. Also, GMPilot, like any other LLM, is susceptible to producing plausible but false or hallucinated content, particularly where there is no direct coverage in the knowledge base. This limitation highlights the significance of human confirmation in important decisions.

Lastly, organizational change management is required to achieve successful deployment. The quality professionals should be taught how to formulate effective questions and incorporate the agent in their decision-making process without excessive dependence. The system is not a substitute to human judgment but rather an augmentative tool.

## 6. Holistic Value of GMPilot

This part will focus on the change of mind set between enterprise application and the growth of the pharmaceutical industry, as well as the possible positive effects that may be achieved with the massive implementation and introduction of GMPilot.

GMPilot can create a more consistent and accurate perception of the cGMP compliance requirements in various enterprises and manufacturing facilities by anchoring itself on an authoritative and unified knowledge base. This will reduce the gaps between companies in terms of compliance capabilities that are due to resource differences, raise the overall quality of compliance in the industry, and offer a better foundation of patient safety and the global supply chain sustainability. GMPilot is capable of making the work of compliance much more efficient by freeing quality experts of repetitive high intensity information retrieval and document organization activities. This enables valuable human and financial resources to be reoriented to more valuable activities like process innovation and enhancement of quality systems, maximization of allocation of resources within the industry.

The knowledge base of GMPilot is, in effect, a constantly updated and structured database of compliance experience. The cases, interpretations, and best practices included in the system will continue to enrich as its knowledge base increases. This gives an effective platform of accelerated learning and avoiding common pitfalls to the pharmaceutical industry particularly small and medium sized enterprises and those in emerging markets which have been the pain point of compliance knowledge loss because of personnel turnover. In case of any changes in regulations, the knowledge base of GMPilot can be relatively fast to incorporate the new rules and distribute the revised guidance to all users through the platform. This feature allows the whole industry to become more responsive to regulatory changes, making it stronger against the growing complexity of the regulatory framework.

## 7. Conclusion

This paper has presented GMPilot, a specialized AI agent that is aimed at fulfilling the strict requirements of the current Good Manufacturing Practice imposed by the FDA as an intelligent quality expert in the 24/7 quality management process. This paper shows how it is possible to implement GMPilot and offer highly context-sensitive and traceable professional support to quality professionals through building an authoritative regulatory knowledge system and integrating it with an agent architecture based on ReAct and RAG technologies. Its fundamental contribution is to suggest and pre-validate an AI agent paradigm of high-regulated pharmaceutical fields (e.g., FDA cGMP). This paradigm guarantees basic enhancements in output precision, traceability and response effectiveness, with deep domain personalization, high-fidelity knowledge base and adaptive ReAct workflow. GMPilot represents an evolutionary journey into the methods of quality management where the focus is no longer on relying on individual skills and fragmented documents but instead on a constantly integrating and real-time responsive intelligent agent. In this regard, GMPilot enables professionals to make more effective, consistent and evidence-based decisions, which eventually enhances corporate quality assurance capabilities, risk reduction and provision of safer and high-quality products to patients.

Our main point is that the future of compliance is not about automating manual processes but rather rethinking the entire workflow around continuous and intelligent support. GMPilot is an active search towards this future, with the goal of becoming an indispensable AI expert partner for pharmaceutical quality departments.

## 8. Future work

Further development would still proceed to optimize the useability and performance of GMPilot. Future work will also explore expansion to additional regulatory jurisdictions and further evaluation of human-AI collaboration effectiveness in real-world quality operations.

In conclusion, continuous development of compliance support AI agent, such as GMPilot, will contribute to the creation of a culture of quality excellence and proactive compliance that will bring long-term benefits to both the industry and public health.

**Acknowledgement:**

This project is supported by Youmin Wang (Silver Spring Biopharma Consulting LLC), thanks for Wang's great work and support.